\documentclass{Interspeech}
\usepackage{multirow}
\usepackage{subcaption}
\usepackage{parskip}
\usepackage{xurl}
\usepackage{hyperref}

\interspeechcameraready

\title{Semantic-Aware Interpretable Multimodal Music Auto-Tagging}

\author[affiliation={1}]{Andreas}{Patakis}
\author[affiliation={1}]{Vassilis}{Lyberatos}
\author[affiliation={1}]{Spyridon}{Kantarelis}
\author[affiliation={1}]{Edmund}{Dervakos}
\author[affiliation={1}]{Giorgos}{Stamou}

\affiliation{National Technical University of Athens}{Athens}{Greece}
\email{andreaspatakis@mail.ntua.gr, \{vaslyb, spyroskanta, eddiedervakos\}@ails.ece.ntua.gr, gstam@cs.ntua.gr}

\usepackage{comment}

\begin{document}
\small

\maketitle

\begin{abstract}

Music auto-tagging is essential for organizing and discovering music in extensive digital libraries. While foundation models achieve exceptional performance in this domain, their outputs often lack interpretability, limiting trust and usability for researchers and end-users alike. In this work, we present an interpretable framework for music auto-tagging that leverages groups of musically meaningful multimodal features, derived from signal processing, deep learning, ontology engineering, and natural language processing. To enhance interpretability, we cluster features semantically and employ an expectation maximization algorithm, assigning distinct weights to each group based on its contribution to the tagging process. Our method achieves competitive tagging performance while offering a deeper understanding of the decision-making process, paving the way for more transparent and user-centric music tagging systems.

\end{abstract}

\section{Introduction}\label{sec:introduction}

In the era of expansive digital music repositories, automatic music auto-tagging has become essential for improving discoverability, personalization, and organization. This task involves assigning semantic labels—such as genres, instruments, moods, and production qualities—to audio tracks. Over the years, advances in deep learning, and specifically the foundation models in music~\cite{ma2024foundation}, have significantly improved tagging accuracy, leveraging large-scale datasets and powerful models \cite{li2023mert,lee2019sample,won2024foundation}. However, the complex and opaque nature of these models raises concerns about interpretability. Users and researchers often lack insights into how specific tags are assigned, which undermines trust and limits the utility of these systems in scenarios requiring human-aligned explanations \cite{doshi2017interpretability}.

Interpretability in music tagging is essential due to the subjective and multidimensional nature of music perception. To tackle this challenge, researchers have explored various approaches, including attention mechanisms~\cite{selfattentionmusicinterpretability}, post-hoc explanations~\cite{sotirou2024musiclime}, concept-based explanations~\cite{foscarin2022concept}, and prototypes grounded in genres or spectral patterns~\cite{alonso2024leveraging, loiseau22amodelyoucanhear}. Despite their potential, these methods often rely on explaining black-box models, which can produce misleading interpretations~\cite{rudin2019stop}.

A more reliable solution is to design interpretable models from the ground up and train them on well-defined, human-understandable features. Perceptual musical features serve as ideal candidates for this purpose, as they are inherently comprehensible to human users~\cite{aljanaki2018data,chowdhury2019towards,lyberatos2024perceptual}. These features encompass both musical and acoustic elements—such as tonality, rhythmic stability, and loudness—and are strongly linked to emotional responses~\cite{gabrielsson2010role,wedin1972multidimensional}. However, existing research often neglects the inherently multimodal nature of music. Previous studies have shown that lyrical features can be indicative of different genres \cite{mayer2008rhyme,fell2014lyrics}. To bridge this gap, we propose an interpretable music auto-tagging pipeline that integrates perceptual features derived from both audio and lyrics, enabling a more holistic and transparent understanding of music tagging.

We focus on extracting features that are not only predictive but also understandable to musicologists, practitioners, and end-users. These features are derived from diverse approaches: signal processing captures low-level audio characteristics \cite{bogdanov2013essentia}; DNNs extract mid-level features \cite{chowdhury2019towards}; ontology engineering extract harmonic features \cite{kantarelis2022functional}, and NLP methods leverage lexical, rhythmical, and phonological lyrical features.

Additionally, previous works leveraging perceptual features often rely on algorithms with  uncertainty, such as \textsc{XGBoost} \cite{lyberatos2024perceptual}. To enhance interpretability, we employ an Expectation Maximization (EM) - based algorithm to cluster extracted features based on their semantic similarity. This approach, which has provided valuable insights in computational neuroscience~\cite{fuglsang2024exploring}, enables us to assign distinct weights to each group, reflecting their contribution to specific tagging tasks. By structuring the feature space in this way, our framework helps users discern the relative importance of different musical attributes in tag predictions.

The contributions can be succinctly summarized as follows: 
\begin{itemize} 
    \item Provided clear and certain global explanations for the music auto-tagging process.
    \item Introduced the incorporation of features drawn from song lyrics alongside audio-based features, making the approach both explainable and multimodal.
    \item Explored three distinct semantic grouping approaches, each designed to provide unique explanations and insights tailored to different contexts.
\end{itemize}

\begin{figure*}[!t]
    \centering
    \includegraphics[width=0.83\linewidth]{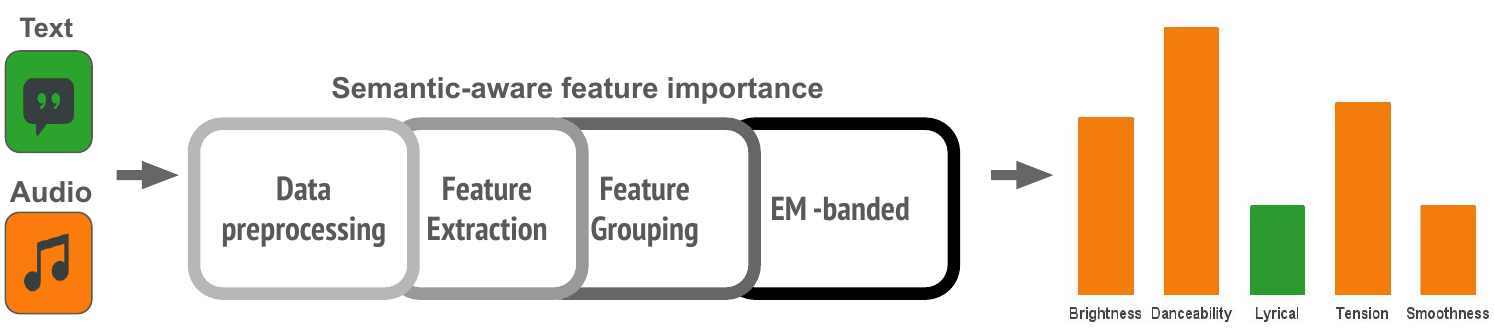}
    \caption{Overview of the pipeline of our methodology.}
    \label{fig:pipeline}
    \vspace{-10pt}
\end{figure*}

\section{Methodology}\label{sec:features}

    The methodology begins with feature extraction, followed by the grouping of related features to support the model's understanding~\footnote{For more details on features and groupings, see\par {\fontsize{6.3pt}{7pt}\selectfont{\url{https://github.com/AndreasPatakis/SAMAT/blob/main/Supp.pdf}}}}. Finally, the \textsc{EM-banded} algorithm processes these grouped features to learn, analyze, and generate predictions, while simultaneously providing semantic explanations of its operation. An overview of the methodology is provided in Figure~\ref{fig:pipeline}. 

\subsection{Extracting Perceptual Multimodal Features}\label{sec:extract}

Building upon the pipeline proposed in~\cite{lyberatos2024perceptual}, we extracted perceptual features using four distinct techniques: deep learning models, ontology engineering, signal processing, and natural language processing.

We used \textbf{symbolic knowledge} to extract harmonic features based on music theory and harmony, building semantic graph databases for each dataset to extract features from track chords. We selected Omnizart \cite{wu2021omnizart} for chord recognition. The databases represent chords as instances, categorized by the Functional Harmony Ontology (FHO)~\cite{kantarelis2022functional}, detailing each chord's role in the track's harmonic structure. We then calculated the ratio of dominant and subdominant chords, creating 6 harmonic features: \textit{Dominants}, \textit{Subdominants}, and specific types such as \textit{MajorDominants}, \textit{MixolydianSeventhDominants}, \textit{AeolianSeventhDominants}, and \textit{MinorFourthSubdominants}. Based on these features we also extracted bigrams and trigrams to use as complementary features. We ended up with 312 n-grams in total.

We harnessed the capabilities of \textbf{deep neural networks} (DNNs) to efficiently extract perceptual features from a dataset curated by music specialists~\cite{aljanaki2018data}. Our DNN model was built on a VGG-ish architecture. We began by extracting 40 Mel-frequency cepstral coefficients (MFCCs) from the first 15 seconds of each audio file, using a window and FFT size of 2048. The model was then trained on the dataset as a regression task to predict 7 scalar perceptual features: \textit{Melodiousness}, \textit{Articulation}, \textit{Rhythmic Stability}, \textit{Rhythmic Complexity}, \textit{Dissonance}, \textit{Tonal Stability}, and \textit{Minorness}.

By applying \textbf{signal processing}, we extracted acoustically meaningful features. To ensure that the selected features are appropriate for our task, a filtering process was necessary, with interpretability as the key criterion.
For this task, we utilized a signal processing tool called Essentia\footnote{\fontsize{6.4pt}{7pt}\selectfont{\url{https://essentia.upf.edu/streaming_extractor_music.html}}} \cite{bogdanov2013essentia} to extract the features. 
The final selection consisted of 85 scalar features, including: \textit{Danceability}, \textit{Loudness}, \textit{Chords Changes Rate}, \textit{Dynamic Complexity}, \textit{Zero Crossing Rate}, \textit{Chords Number Rate}, \textit{Pitch Salience}, \textit{Onset Rate}, \textit{BPM}, \textit{Spectral Energybands}, in combination with \textit{Pulse}, \textit{Attack}, \textit{Spectral}, \textit{MFCCs} and \textit{Chroma} features.

We employed \textbf{natural language processing} (NLP) techniques to extract phonological, lexical, and rhythmic features from the lyrics.
For datasets without song lyrics, we incorporated a source-separation step using Demucs~\cite{rouard2023hybrid} to isolate the vocal track before transcription and then retrieved them by using Whisper~\cite{radford2023robust}, an automatic speech recognition system, to transcribe lyrics directly from the audio. 
We ended up with 9 scalar features extracted through NLP. These features are \textit{Alliteration}, \textit{Assonance}, \textit{Consonance}, \textit{Rhyme Density}, \textit{Syllables per Word}, \textit{Total Syllables}, \textit{Word Repetition Rate}, \textit{Lexical Diversity} and \textit{Words per Second}.

\subsection{Semantic Grouping of the Features}\label{sec:semantic}

\begin{table*}[h]
    \centering
    \resizebox{0.6\textwidth}{!}{
    \begin{tabular}{l l cc cc cc}
        \toprule
        & & \multicolumn{2}{c}{\textsl{MTG-Jamendo}} & \multicolumn{2}{c}{\textsl{Music4All}} & \multicolumn{2}{c}{\textsl{AudioSet}}\\
        \cmidrule(lr){3-4} \cmidrule(lr){5-6} \cmidrule(lr){7-8}
        Approach & Model & w/ & w/o & w/ & w/o & w/ & w/o\\
        \midrule
        & Multimodal  & 75.90 & 69.74     & 57.34 & 53.75 & 58.71 & 52.10\\
        \midrule
        \multirow{2}{*}{User-Friendly} 
        & \textsc{EM-banded}   & 75.75 & 74.81 & 43.18 & 40.66 & 37.04 & 37.04\\
        & \textsc{XGBoost}     & 73.87 & 72.27 & 46.81 & 44.92 & 46.03 & 33.33\\
        \midrule
        \multirow{2}{*}{Domain-Expert} 
        & \textsc{EM-banded}   & 73.01 & 71.63 & 40.72 & 38.25 & 35.19 & 33.33\\
        & \textsc{XGBoost}     & 72.88 & 69.54 & 45.77 & 42.73 & 50.00 & 31.48 \\
        \midrule
        \multirow{2}{*}{All-Features}  
        & \textsc{EM-banded}   & 76.95 & 76.61 & 45.09 & 43.45 & 44.44 & 48.15\\
        & \textsc{XGBoost}     & 74.64 & 73.29 & 48.59 & 46.36 & 48.15 & 38.89 \\
        \bottomrule
    \end{tabular}
    }
    \caption{Performance results for all approaches, with and without lyrics; ROC-AUC for MTG-Jamendo, accuracy for all other datasets.}
    \label{tab:combined_all_approaches}
    \vspace{-15pt}
\end{table*}

To obtain meaningful feature groupings, we organized the features into three semantic categories: two based on heuristic categorization by music experts and the other on their source of origin. Using these structured feature groups, we trained the \textsc{EM-banded} algorithm (see Section~\ref{subsec:embanded}) for music auto-tagging, selecting it for its ability to leverage the inherent structure of the features. 

The first category, namely \textit{User-Friendly}, consists of groups that may be more intuitive for listeners familiar with common music terms. Specifically, we grouped together features that are not only conceptually related but also distinct enough to be easily distinguished by most listeners, as they do not significantly overlap. Additionally, these features share the characteristic that higher values correspond to similar perceptual attributes. This resulted in five groups: \textit{Brightness \& Sharpness}, \textit{Danceability \& Rhythm}, \textit{Tension \& Complexity}, \textit{Acoustic Smoothness} and \textit{Lyrical}. For example, all features within the \textit{Brightness \& Sharpness} group indicate that higher values correspond to a brighter and sharper sound in the music track.  The total number of features used in this approach was 67.

The second category, \textit{Domain-Expert}, targets individuals with music or production expertise, grouping features based on musical dimensions without requiring consistent value interpretation. For instance, \textit{Dissonance} and \textit{Tonal Stability} are grouped despite opposite value orientations. In this approach, we ended up with five groups: \textit{Spectral Features} (features related to the spectral characteristics of a music track), \textit{Harmonic Features} (features describing harmonic and melodic properties), \textit{Rhythmic Features} (features capturing rhythmic elements), \textit{Sound Shaping Features} (features related to the timbral and dynamic characteristics of the sound) and \textsl{Lyrical}. The total number of features used in this approach was 42. Though there is some overlap between the \textit{User-Friendly} and \textit{Domain-Expert} categories, the distinction lies in their focus. User-Friendly groups are based on intuitive terms, while Domain-Expert groups focus on technical aspects requiring specialized knowledge. For example, \textit{Brightness} refers to perception, while \textit{Spectral} features involve frequency analysis. The \textit{Domain-Expert} approach has fewer groups due to its specialization, while \textit{User-Friendly} covers a broader range.

The third and final category, \textit{All-Features}, consists of using all the available features, and grouping them based on the method of extraction. Four groups were created in total, aligning with the four methods of feature extraction presented in Section~\ref{sec:extract}. Even though these groups are not as comprehensive as the ones in previous approaches, including an approach where all the features are used can help us understand the full capabilities of the models used.

\subsection{\textsc{EM-banded}}\label{subsec:embanded}
While accurate predictions are important in music auto-tagging, gaining insight into which types of features contribute to specific tags can enhance interpretability. This motivates the exploration of approaches that enable group-wise feature analysis to better understand their roles in tagging decisions. To address this, we adopt the \textsc{EM-banded} model proposed in \cite{fuglsang2024exploring}. As stated in the original paper, regression analysis is a key method in domains like computational neuroscience for modeling where a common approach involves fitting linear models with predictors organized into distinct groups, such as subsets of stimulus features in encoding models or neural response channels in decoding models. 

By using the expectation-maximization (EM) algorithm, the model tunes the hyperparameters that dictate the variances of the prior distributions, enabling differential shrinkage for specific sets of regression weights. The regression model considered is in the following form:
\vspace{-0.6em}
\begin{align}
    y = F_1 \beta_1 + F_2 \beta_2 + ... + F_J \beta_J + \epsilon
\label{eq:em_regression_form}
\end{align}

where $\beta_j$ denotes regression weights for a given group of predictors, $F_j$.
\vspace{-0.6em}

A zero-mean Gaussian prior distribution is placed over the weights:
\vspace{-0.6em}
\begin{align}
    p(w|\Lambda, \eta, \tau ) = N (w|0, \Lambda)
\label{eq:norm_w}
\end{align}
where $\Lambda$ is a $D x D$ block-diagonal covariance matrix. 

It is important to recognize that the above problem formulation inherently enables feature selection. This is due to the prior distribution over the weights, which effectively shrinks the estimated regression coefficients. The covariance matrix $\Lambda$ of the weights is assumed to follow a block structure, defined as follows:
\vspace{-0.3em}
\begin{align}
\Lambda \equiv \begin{pmatrix}
\lambda_1I_{D_1}  & \cdots & 0 \\
\vdots & \ddots & \vdots \\
0 & \cdots  & \lambda_jI_{D_j}
\end{pmatrix}
\begin{pmatrix}
\Omega_1  & \cdots & 0 \\
\vdots & \ddots & \vdots \\
0 & \cdots  & \Omega_j
\label{mat:matrix_lambda}
\end{pmatrix} 
\end{align}

The authors specify that each of the $j = 1, ..., D_j$ blocks in $\Lambda$ corresponds to a distinct group of predictors, $F_j$. Instead of using a single shared hyperparameter—as in Ridge regression—separate hyperparameters $\lambda_j$ are assigned to each group, allowing for group-specific regularization. Furthermore, Inverse-Gamma prior distributions are placed over the $\lambda_j$ terms, which are parameterized by $\eta$ and $\tau$.

Our approach uses the values of the lambda coefficients to assess the importance of each feature group in the auto-tagging task. Since each group has its own lambda controlling the variance of its coefficients, the magnitude of lambda serves as a direct indicator of that group's relevance to the model.

\section{Experiments}

\subsection{Experimental Setup}

We conducted our experiments using three multimodal music datasets that include both audio and lyrics: two curated from \textsl{Audioset} and \textsl{Music4All}, as described in~\cite{sotirou2024musiclime}, resulting in single-label datasets; and the \textsl{MTG Jamendo} dataset~\cite{bogdanov2019mtg}, which required preprocessing to extract lyrics (see Section~\ref{sec:extract}).
\textsl{Music4All} encompasses 60,000 songs with metadata (artist, album, release year, language) and user-generated tags, while \textsl{AudioSet} contains 308 audio-lyrics pairs with descriptive labels (e.g., \textit{fireworks}, \textit{harmonica}) from YouTube segments. \textsl{MTG-Jamendo} is an open-access collection of 18,486 tracks with 56 mood/theme tags. For evaluation, we used accuracy metrics for the multi-class classification tasks in \textit{Music4All} and \textit{AudioSet}, while AUC-ROC was applied to the multi-label \textit{MTG-Jamendo} dataset.

\begin{figure*}[h]
    \centering
    \includegraphics[width=0.7\linewidth]{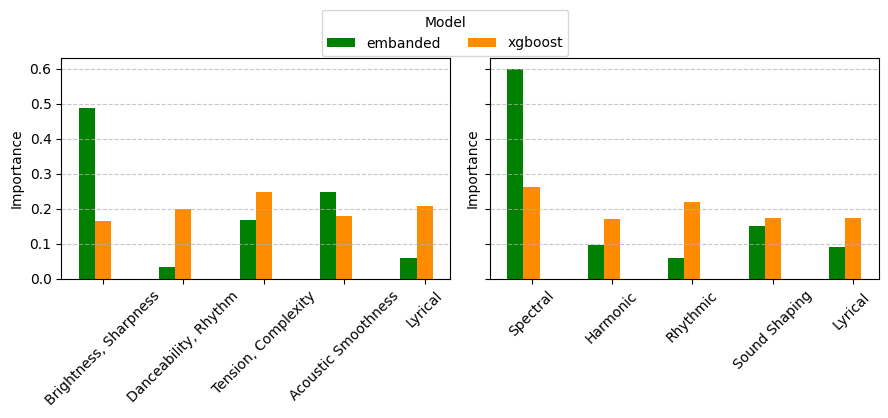}
    \caption{Group importances for \textsl{Music4All} for User-Friendly and Domain-Expert approaches.}
    \label{fig:m4a_importances}
    \vspace{-10pt}
\end{figure*}

After extracting the perceptual features (see Section~\ref{sec:extract}), a series of experiments were conducted. The \textsc{EM-banded} algorithm was applied to evaluate four different feature groupings: two based on semantic categories, one derived from the feature extraction methodology  (see Section~\ref{sec:semantic}) and one where features were randomly assigned to five artificial groups. In this study, the performance and group-wise descriptive capacity of both the \textsc{EM-banded} and \textsc{XGBoost} algorithms were assessed, with comparisons made to the state-of-the-art transformer-based model introduced in \cite{sotirou2024musiclime}, referred to as \textsc{Multimodal}. Additionally, a human evaluation\footnote{\raggedright \fontsize{6.4pt}{7pt}\selectfont{\url{https://forms.gle/3LuZK5CqKUXbfWGj6}}} was conducted to assess the explanations provided. All experiments were implemented in Python~\footnote{\fontsize{6.4pt}{7pt}\selectfont{\url{https://github.com/AndreasPatakis/SAMAT}}} and executed on a system equipped with a GeForce GTX 1080 GPU with 8 GB of VRAM, as needed.

\subsection{Results}
We evaluated the \textsc{EM-banded} and \textsc{XGBoost} models over five runs using randomly sampled training sets. The \textsc{EM-banded} model exhibited no variation in performance, while \textsc{XGBoost} varied by at most 0.5\%. Given this minimal variance, we report the mean performance and consider the results to be stable.

Regarding models' effectiveness,
Table \ref{tab:combined_all_approaches} shows that \textsc{EM-banded} achieves the highest performance (76.95\%) on \textsl{MTG-Jamendo}, surpassing both \textsc{Multimodal} (75.90\%) and \textsc{XGBoost} (74.64\%). On \textsl{Music4All} and \textsl{AudioSet}, \textsc{Multimodal} performs best (57.34\% and 58.71\% respectively), while \textsc{EM-banded} remains competitive (45.09\% and 48.15\%). 
Lyrical features consistently improve performance in almost all models, highlighting their importance in music classification. Interestingly, \textsc{EM-banded} with \textit{Domain-Expert} features slightly outperforms its all-features variant on \textsl{Music4all}, suggesting that carefully selected features can be more effective than using all available features, particularly when interpretability is prioritized.

\begin{table}[!b]
\vspace{-8pt}
    \centering
    \resizebox{0.45\textwidth}{!}{
    \begin{tabular}{lrrr}
    \toprule
    Groups & \textsc{EM-banded} & \textsc{XGBoost} & Human Evaluation \\
    \midrule
    Brightness, Sharpness & 1 & 5 & 3 \\
    Danceability, Rhythm & 5 & 4 & 1 \\
    Tension, Complexity & 3 & 1 & 4 \\
    Acoustic Smoothness & 2 & 3 & 2 \\
    Lyrical & 4 & 2 & 5 \\
    \midrule
    \textbf{Absolute Difference} & 8 & 12 & 0 \\
    \bottomrule
    \end{tabular}
    }
    \caption{Group ranking per algorithm, with Absolute Difference showing the total distance from Human Evaluation results.}
    \label{tab:my_label}
\end{table}

As far as the grouping of feature importance is concerned, Figure~\ref{fig:m4a_importances} demonstrates \textsc{EM-banded}'s superior ability to distinguish relevant patterns based on semantic similarity, revealing clear attribution differences compared to \textsc{XGBoost}'s more uniform distribution. This aligns with established research on genre classification \cite{4895319}, where spectral features consistently emerge as key discriminators. The coherent importance patterns shown by \textsc{EM-banded} particularly validate its effectiveness in capturing these well-documented spectral relationships.

To assess the robustness of each model’s group-level explanations, we evaluated the importance of features randomly assigned to the five artificial groups. Both \textsc{EM-banded} and \textsc{XGBoost} showed relatively uniform distributions of importance ($\approx$ 0.2) across the synthetic groupings, aligning with the expectation that no meaningful structure exists in random partitions. Additionally, we evaluated each algorithm’s group-level rankings using the \textit{User-Friendly} grouping of \textsl{Music4All}, selected for its size and representativeness, with feedback provided by 10 amateur musicians, as shown in Table \ref{tab:my_label}. This human evaluation thus offers a real-world perspective on which musical attributes—like brightness, danceability, or tension—are most relevant. Notably, \textsc{EM-banded}’s total distance from the human rankings (8) was lower than \textsc{XGBoost}’s (12), suggesting that \textsc{EM-banded}’s learned group-level importances better align with users' intuition.

\subsection{Discussion}

In the comparative analysis, the \textsc{Multimodal} model typically exhibited the strongest performance overall. Nevertheless, the \textsc{EM-banded} model consistently delivered effective results in various scenarios, and the \textsc{XGBoost} model matched the performance of the \textsc{EM-banded} model, even surpassing it in the \textsl{AudioSet} dataset.
The key advantage, though, of \textsc{EM-banded} algorithm lies in its native, group-centric interpretability, which is especially evident when using the \textit{User-Friendly} grouping of features. In this setup, features are divided into intuitive categories—like rhythm-related attributes, melodic components, or lyrical topics—so that even non-technical users can easily understand how different aspects of the music contribute to genre classification. This contrasts with the transformer-based \textsc{Multimodal} model, which offers limited transparency into its decision-making process, and with \textsc{XGBoost}, where per-feature importance must be manually aggregated to extract group-level insights—by computing "gain" variable importances, averaging them per group, and normalizing the group importances to sum to 1—a process that inherently introduces a degree of ambiguity into its decisions.

\section{Conclusion \& Future Work}

An important contribution of this work is demonstrating how feature grouping—whether \textit{User-Friendly}, \textit{Domain-Expert}, or otherwise—can be integrated seamlessly into a music tagging pipeline to deliver clear, high-level explanations. By leveraging our approach of \textsc{EM-banded} model we achieve to not only preserve good predictive performance but also provide interpretable, deterministic group-level importance scores that resonate with both expert and non-expert users. This capability is particularly relevant for real-world applications, where understanding why certain features (e.g., rhythmic patterns, spectral descriptors) influence genre predictions is crucial. In addition, the flexibility to choose or customize groupings—for instance, reflecting specific research questions or domain constraints—further enhances the model’s adaptability. Future work will explore refined grouping strategies, aiming to capture a broader range of musical attributes and extend the explainability framework to more diverse contexts in music analysis.

\section{Acknowledgments}
The research project is implemented in the framework of H.F.R.I call “Basic research Financing (Horizontal support of all Sciences)” under the National Recovery and Resilience Plan “Greece 2.0” funded by the European Union –NextGenerationEU(H.F.R.I. Project Number: 15111 - Emotional Artificial Intelligence in Music Expression).

\bibliographystyle{IEEEtran}
{\footnotesize
\bibliography{main}}
\appendix

\end{document}